\documentclass{bmvc2k}
\usepackage{multirow, booktabs, tabularx}
\usepackage{amssymb}
\usepackage{amsmath,amsfonts}

\usepackage{siunitx}
\usepackage{soul}

\usepackage[font=small,labelfont=bf,figurename=Fig.,tablename=Tab.]{caption}
\usepackage{subcaption}
\usepackage[nameinlink]{cleveref}
\crefname{figure}{Fig.}{Figs.}
\crefname{table}{Tab.}{Tabs.}
\crefname{section}{Sec.}{Secs.}
\crefname{subsection}{Sec.}{Secs.}

\usepackage{fixmath}
\renewcommand{\vec}[1]{\mathbold{#1}}

\newcommand{\txt}{\mathbold{r}}
\newcommand{\txtFeat}{\mathbold{p}}

\newcommand{\txtReal}{\mathbold{r^+}}
\newcommand{\txtRealFeat}{\mathbold{p}^+}

\newcommand{\txtWrong}{\mathbold{r}^-}
\newcommand{\txtWrongFeat}{\mathbold{p}^-}

\newcommand{\img}{\mathbold{v}}
\newcommand{\imgFeat}{\mathbold{q}}

\newcommand{\imgReal}{\mathbold{v^+}}
\newcommand{\imgRealFeat}{\mathbold{q}^+}

\newcommand{\imgWrong}{\mathbold{v^-}}
\newcommand{\imgWrongFeat}{\mathbold{q}^-}

\newcommand{\imgFake}{\mathbold{\tilde{v}^+}}
\newcommand{\imgFakeFeat}{\mathbold{\tilde{q}}^+}

\DeclareMathOperator{\R}{\mathbb{R}}
\DeclareMathOperator{\TxtEnc}{F_p}
\DeclareMathOperator{\ImgEnc}{F_q}
\DeclareMathOperator{\Ncal}{\mathcal{N}}
\DeclareMathOperator{\Lcal}{\mathcal{L}}
\DeclareMathOperator*{\softmax}{softmax}

\usepackage{xspace}
\newcommand{\FoodSpace}{\texttt{FoodSpace}\xspace}

\title{The Art of Food: Meal Image Synthesis from Ingredients}

\addauthor{Fangda Han}{fh199@cs.rutgers.edu}{1}
\addauthor{Ricardo Guerrero}{r.guerrero@samsung.com}{2}
\addauthor{Vladimir Pavlovic}{vladimir@cs.rutgers.edu}{1}

\addinstitution{
Rutgers University\\
Piscataway, NJ, USA
}
\addinstitution{
Samsung AI Center\\
Cambridge, UK
}

\runninghead{Under review}{as a conference paper at BMVC 2019}

\def\eg{\emph{e.g}\bmvaOneDot}

\begin{document}

\maketitle

\begin{abstract}
In this work we propose a new computational framework, based on generative deep models, for synthesis of photo-realistic food meal images from textual descriptions of its ingredients.  
Previous works on synthesis of images from text typically rely on pre-trained text models to extract text features, followed by a generative neural networks (GANs) aimed to generate realistic images conditioned on the text features. These works mainly focus on generating spatially compact and well-defined categories of objects, such as birds or flowers.
In contrast, meal images are significantly more complex, consisting of multiple ingredients whose appearance and spatial qualities are further modified by cooking methods.  
We propose a method that first builds an attention-based ingredients-image association model, which is then used to condition a generative neural network tasked with synthesizing meal images.
Furthermore, a cycle-consistent constraint is added to further improve image quality and control appearance.
Extensive experiments show our model is able to generate meal image corresponding to the ingredients, which could be used to augment existing dataset for solving other computational food analysis problems.
\end{abstract}


\section{Introduction}
\label{sec:intro}

Computational food analysis (CFA) has recently drawn substantial attention, in part due to its importance in health and general wellbeing \cite{carvalho2018cross, salvador2017learning, horita2018food, chen2018deep, ahn2011flavor, teng2012recipe, martinel2018wide, aguilar2018grab, DBLP:journals/corr/abs-1808-07202}. 
For instance, being able to extract food information including ingredients and calories from a meal image could help us monitor our daily nutrient intake and manage our diet. 
In addition to food intake logging, CFA can also be crucial for learning and assessing the functional similarity of ingredients, meal preference forecasting, and computational meal preparation and planning \cite{teng2012recipe, helmy2015health}.
Automatically recognizing ingredients in meal images is non-trivial due to complex dependencies between raw ingredients and their appearance in prepared meals. 
For instance, the appearance of a shredded, baked apple differs significantly from that of the fresh whole fruit. 
Ingredients can remain compact during the preparation process or can blend with other food elements, resulting in distributed visual appearance. 
Finally, some ingredients, such as salt, oil, or water, typically leave no footprint in the visible image spectrum, while others, such as blended carrots can nevertheless change the visual appearance by modulating the meal color. 
All of these make the visual identification of ingredients an enormously challenging task.

Recently, deep learning models have brought high accuracy to general image recognition tasks \cite{krizhevsky2012imagenet, he2016deep}.
To achieve this, such models require large amounts of structured and labeled data, often collected and processed from web resources.  
However, the recipe information contained within them (\eg images, title, instructions and ingredients names) may be noisy or altogether missing, resulting in weakly structured data~\cite{teng2012recipe, salvador2017learning} of limited utility for deep learning-based CFA.  
Despite this, there has been limited yet encouraging success in visual identification of food ingredients \cite{chen2016deep, bolanos2017food}.

One solution is using synthesized images with specific ingredients, cutting and cooking methods to provide more structured data.
However, the work on generating photo-realistic meal images has so far failed to materialize due to the complex nature of meals and their visual appearance. Among few works,
\cite{Horita:2018:FCT:3230519.3230597} extends CycleGAN \cite{zhu2017unpaired} to transfer food images between ten categories.
Nonetheless, this work is limited to image "style" transformation between two types of food without changing the food composition. 
The focus of our work is
to build a model that could generate images from a set of specific ingredients. To the best of our knowledge, this is the first work that attempts 
this task.
%
Much like ingredient identification, meal image synthesis is a nontrivial task, stemming from the same factors such as the appearance diversity, dependency on the cooking method, variations in preparation style, visual dissolution (\eg, salt or butter may have no visual signatures), etc. 
As a consequence, the generative meal model has to infer these key pieces of information implicitly.

In this work, we propose a model to generate synthetic photo-realistic meal image conditioned on a set of ingredients.
The main contributions are: 1) Combining attention-based recipe association model\cite{chen2018deep} and StackGAN \cite{zhang2017stackgan++} to generate meal images from ingredients. 2) Adding a cycle-consistency constraint to further improve image quality and control the appearance of the image by changing ingredients.
%

\section{Related Work}

Generative neural networks (GAN) are a popular type of generative models for image synthesis \cite{goodfellow2014generative}. 
In general, vanilla GANs learn to model the distribution of real images via a combination of two networks, one that generates images from a random input vector and another that attempts to discriminate between real and generated images.

Work on generating images conditioned on a deterministic label by directly concatenating the label with the input was proposed by~\cite{mirza2014conditional} and by adding the label information at a certain layer's output in~\cite{odena2017conditional, miyato2018cgans}. 
Another line of work conditions the generation process with text information.
\cite{reed2016generative} uses a pre-trained model to extract text features and concatenate them with the input random vector, in order to generate text-based images.
\cite{zhang2017stackgan++} extends this concept by stacking three GANs to generate the same image at different resolutions.
The same idea was shown to be valuable in human face generation \cite{karras2017progressive, karras2018style}. 
%
These works are conditioned on short textual descriptions of the image and rely on recurrent neural networks (RNN) to extract text features. RNNs treat words sequentially and with the same importance, in the sparse set of ingredients of a meal, not all ingredients play an important role in image appearance, therefore, it makes sense to model this importance. 
Inspired by \cite{chen2018deep}, we combine attention mechanism with bi-directional LSTM to learn the importance of each ingredient, the attention-based LSTM model improves the association performance considerably.

More critically, most prior works implicitly assume the visual categories modeled, such as birds, flowers, or faces, are well-structured singular objects, consistent in appearance.  Meal images, on the other hand, have more variable appearance when conditioned on ingredients.  An important difference between StackGAN and the work proposed here is the addition of a cycle-consistency constraint to address this challenge.
The intuition is that if the generated image is of high quality and captures the ingredients correctly, it should extract similar feature as that from the real image. Experiment shows this constraint could improve the image quality and prevent mode collapse.

\section{Methodology}


\subsection{Model Structures}
\label{sec:model_structures}

To generate meal images from ingredients, we first train a recipe association model to find a shared latent space between ingredient sets and images, and then use the latent representation of ingredients to train a GAN to synthesize meal images conditioned on those ingredients.

\subsubsection{Attention-based Cross-Modal Association Model in \FoodSpace}
\label{subsec:association_model}

\begin{figure*}[ht]
\begin{center}
\includegraphics[width=\textwidth]{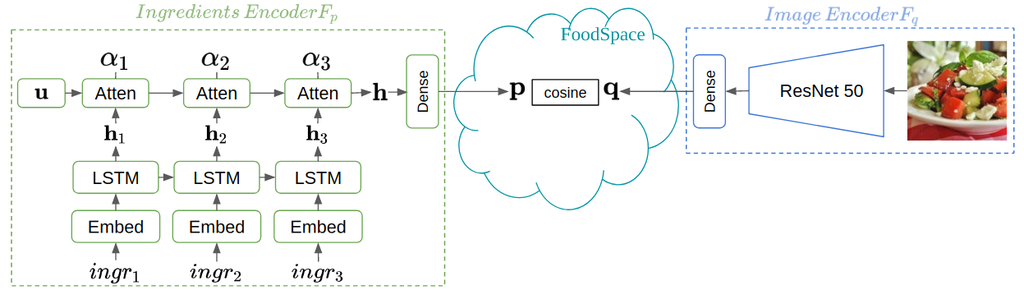}
\end{center}
\caption{The framework of the attention-based recipe association model.}
\label{fig:framework_association}
\end{figure*}

In order to extract ingredients feature, we use an attention based cross-modal association model \cite{chen2018deep} trained to match ingredient sets and their corresponding images in a joint latent space, denoted the \FoodSpace.
During training, the model takes a triplet as input, the recipe ingredients, its corresponding image, and an image from another recipe, $(\txtReal, \imgReal, \imgWrong)$, respectively. Using two separate neural networks, one for ingredients, $\TxtEnc$ and another for images, $\ImgEnc$, the triplet is embedded in the \FoodSpace with coordinates $(\txtRealFeat, \imgRealFeat, \imgWrongFeat)$. The networks are trained to maximize the association in \FoodSpace between positive pair $(\txtRealFeat,\imgRealFeat)$, at the same minimizing the association between negative pair $(\txtRealFeat,\imgWrongFeat)$.

Formally, with the ingredients encoder $\txtFeat = \TxtEnc(\txt)$ and image encoder $\imgFeat = \ImgEnc(\img)$, the training is a maximization of the following objective function, 
\begin{align}
\begin{split}
    V(\TxtEnc, \ImgEnc) = 
    & \mathbb{E}_{ \hat{p}(\txtReal,\imgReal), \hat{p}(\imgWrong) } \min\left( \left[ \cos{\left[\txtRealFeat, \imgRealFeat\right]} - \cos{\left[\txtRealFeat, \imgWrongFeat\right]} - \epsilon \right], 0 \right) + \\
    & \mathbb{E}_{ \hat{p}(\txtReal,\imgReal), \hat{p}(\txtWrong) } \min\left( \left[ \cos{\left[\txtRealFeat, \imgRealFeat\right]} - \cos{\left[\txtWrongFeat, \imgRealFeat\right]} - \epsilon \right], 0 \right),
\end{split}
\end{align}
where $\cos{\left[\txtFeat, \imgFeat\right]} = \txtFeat^\intercal\imgFeat / \sqrt{ (\txtFeat^\intercal\txtFeat) (\imgFeat^\intercal\imgFeat)}$ is the cosine similarity in \FoodSpace and $\hat{p}$ denote the corresponding empirical densities on the training set. We combine the cosine similarity of the positive pair and that of the negative pair together and we add a margin $\epsilon$ to make the model focus on those pairs that are not correctly embedded, we empirically set it to $0.3$ by cross-validation.
\cref{fig:framework_association} shows a diagram of the attention-based association model. The details of ingredients encoder $\TxtEnc$ and image encoder $\ImgEnc$ are explained below.

\paragraph{Ingredients encoder:} $\TxtEnc$ takes the recipe's ingredients as input and outputs their feature representation in the shared space. The goal is to find the ingredient embeddings that reflect dependencies between ingredients, in order to facilitate implicit associations even when some ingredients are not visually observable. 
For this, the model first embeds the one-hot vector of each ingredient into a low-dimension vector ($\R^{300}$),  treating this as a sequence input of a bi-directional LSTM\footnote{Hence, we assume a chain graph can approximate arbitrary ingredient dependencies within a recipe.}.
Instead of using the output of the last layer as the output of the LSTM, each hidden state $\vec{h}_i \in \R^{300}$ is used as the feature of the corresponding ingredient.
As not all ingredients play equally important role in image appearance, we apply an attention mechanism to model the contributions of each ingredient.
During training, the model learns a shared contextual vector $\vec{u} \in \R^{300}$ of the same dimension as the hidden state, used to assess the attention of each ingredient,
\begin{align}
    \left\{\alpha_i\right\} = \softmax\left\{ \vec{u}^T \cdot \vec{h}_i \right\}, \;i\in[1,N],
\end{align}
where $N$ is the number of ingredients in the recipe. The attention-based output of LSTM is a weighted summation of all hidden states, $\vec{h} = \sum_{i=1}^{N} \alpha_i \vec{h}_i$.
Finally, $\vec{h}$ is projected to the shared space to yield the ingredients feature $\txtFeat \in \R^{1024}$ in \FoodSpace. 

\paragraph{Image encoder:} $\ImgEnc$ takes a meal image as input and outputs a feature representing the image in \FoodSpace. Resnet50 \cite{he2016deep} pre-trained on ImageNet is applied as the base model for feature extraction.
In order to get a more meaningful feature of the image, we follow \cite{chen2018deep} and finetune the network on UPMC-Food-101 \cite{wang2015recipe}, we use the activation after the average pooling ($\R^{2048}$) and project it to \FoodSpace to get $\imgFeat \in \R^{1024}$.

\subsubsection{Generative Meal Image Network}
\begin{figure*}[ht]
\begin{center}
\includegraphics[width=\textwidth]{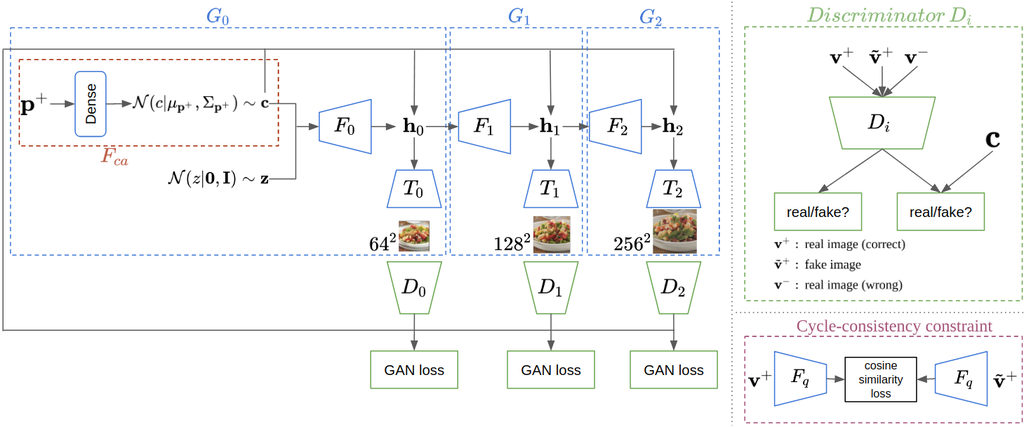}
\end{center}
\caption{Framework of the generative model. $G_0$, $G_1$, $G_2$ represent the branches in generator. $D_0$, $D_1$, $D_2$ represent the discriminators for images of low, medium and high resolution. $F_q$ is the image encoder trained in the association model.}
\label{fig:generator}
\end{figure*}

The generative adversarial network takes the ingredients as input and generates the corresponding meal image.
We build upon StackGAN-v2~\cite{zhang2017stackgan++}, which contains three branches stacked together. Each branch is responsible for generating image at a specific resolution and each branch corresponds to a separate discriminator. 
The intuition is to have three discriminators being responsible for distinguish images at low, medium and high resolutions. The framework is shown in \cref{fig:generator}.

\paragraph{Generator:} The ingredients $\txtReal$ are first encoded using the pre-trained $\TxtEnc$ (fixed during training StackGAN-v2) to obtain text feature $\txtRealFeat$.
Subsequently, $\txtRealFeat$ is forwarded through a conditional augmentation network $F_{ca}$ to estimate the distribution $p(\vec{c}|\txtRealFeat)$ of the ingredient appearance factor $\vec{c}$, modeled as the Gaussian distribution
\begin{align}
    \left(\mu_{\txtRealFeat}, \Sigma_{\txtRealFeat}\right) = F_{ca}(\txtRealFeat), \; \vec{c} \sim p(\vec{c}|\txtRealFeat) = \Ncal(\mu_{\txtRealFeat}, \Sigma_{\txtRealFeat}),
\end{align} 
where $\mu_{\txtRealFeat}$ and $\Sigma_{\txtRealFeat}$ are the mean and the covariance given the ingredients encoding $\txtRealFeat$ in \FoodSpace. 
Intuitively, this sampling process introduces noise to $\txtRealFeat$,  making the model robust to small perturbations in \FoodSpace.
Variational regularization~\cite{kingma2013auto} is applied during training to make $p(c|\txtRealFeat)$ close to the standard Gaussian distribution,
\begin{align}
    \Lcal_{ca} = D_{KL} \left[ \Ncal(\mu_{\txtRealFeat}, \Sigma_{\txtRealFeat}) || \Ncal(\vec{0}, \vec{I})  \right].
\end{align}
Subsequently, $\vec{c}$ is augmented with Gaussian noise $\vec{z} \sim \Ncal(\vec{0}, \vec{I})$ to generate the latent feature $\vec{h}_0 = F_0(\vec{z}, \vec{c})$ for the first branch and the low-resolution image $\imgFake_0  = T_0(\vec{h}_0)$,
where $F_0$ and $T_0$ are modeled by neural networks. Similarly, the medium and high resolution images are generated by utilizing the hidden feature of the previous branches, $\vec{h}_1 = F_1(\vec{h}_0, \vec{c}), \; \imgFake_1 = T_1(\vec{h}_1)$ and $\vec{h}_2 = F_2(\vec{h}_1, \vec{c}), \; \imgFake_2 = T_2(\vec{h}_2)$.
%
Overall, the generator contains three branches, each responsible for generating the image at a specific scale, $G_0 = \{ F_{ca}, F_0, T_0 \} ,\; G_1 = \{ F_1, T_1 \} ,\; G_2 = \{ F_2, T_2 \}$. Optimization of the generator will be described after introducing the discriminators.

\paragraph{Discriminator:} 
Each discriminator's task is three-fold: (1) Classify real, `correctly-paired' $\imgReal$ with ingredient appearance factor $\vec{c}$ as real; (2) Classify real, `wrongly-paired' $\imgWrong$ with $\vec{c}$ as fake; and (3) Classify generated image $\imgFake$ with $\vec{c}$ as fake. 
Formally, we seek to minimize the cross-entropy loss
\begin{align}\small
\label{eq:loss_d_cond}
\begin{split}
    \Lcal^{cond}_{D_i} = 
    & -\mathbb{E}_{\imgReal \sim p_{d_i}} [\log D_i(\imgReal, \vec{c})] 
      + \mathbb{E}_{\imgWrong \sim p_{d_i}} [\log D_i(\imgWrong, \vec{c})] 
     +\mathbb{E}_{\imgFake \sim p_{G_i}} [\log D_i(\imgFake, \vec{c})],
\end{split}
\end{align}
where $p_{d_i}$, $p_{G_i}$, $G_i$ and $D_i$ correspond to the real image distribution, fake image distribution, generator branch, and the discriminator at the $i^{th}$ scale.
To further improve the quality of the generated image, we also minimize the unconditional image distribution as
\begin{align}\small
\label{eq:loss_d_uncond}
\begin{split}
    \Lcal^{uncond}_{D_i} = 
    & -\mathbb{E}_{\imgReal \sim p_{d_i}} [\log D_i(\imgReal)]  
    - \mathbb{E}_{\imgWrong \sim p_{d_i}} [\log D_i(\imgWrong)] 
    + \mathbb{E}_{\imgFake \sim p_{G_i}} [\log D_i(\imgFake)]
\end{split}
\end{align}

\paragraph{Losses:} During training, the generator and discriminators are optimized alternatively by maximizing and minimizing \eqref{eq:loss_d_cond} and \eqref{eq:loss_d_uncond} respectively. All generator branches are trained jointly as are the three discriminators, with final losses
\begin{align}
    \label{eq:l_g}
    \Lcal_G &= \sum_{i=0}^2 \left\{ \Lcal^{cond}_{G_i} + \lambda_{uncond} \Lcal^{uncond}_{G_i}\right\} + \lambda_{ca}\Lcal_{ca}, &
    \Lcal_D &= \sum_{i=0}^2 \left\{ \Lcal^{cond}_{D_i} + \lambda_{uncond} \Lcal^{uncond}_{D_i}\right\},
\end{align}
where $\lambda_{uncond}$ is the weight of the unconditional loss and $\lambda_{ca}$ the weight of the conditional augmentation loss.
We empirically set $\lambda_{uncond}=0.5$ and $\lambda_{ca}=0.02$ by cross-validation.

\subsection{Cycle-consistency constraint}
\label{sec:cycle}
A correctly-generated meal image should "contain" the ingredients it is conditioned on. Thus,
a cycle-consistency term is introduced to keep the fake image contextually similar, in terms of ingredients, to the corresponding real image in \FoodSpace.

Specifically, for a real image $\imgReal$ with \FoodSpace coordinate  $\imgRealFeat$ and the corresponding generated $\imgFake$ with $\imgFakeFeat$, the cycle-consistency regularization aims at maximizing the cosine similarity at different scales, $\Lcal_{C_i} = \cos{\left[\imgRealFeat, \imgFakeFeat\right]}$\footnote{Note that the images in different resolutions need to be rescaled for the input of the image encoder.}.
The final generator loss in \eqref{eq:l_g} becomes 
\begin{align}
    \label{eq:l_g_with_cycle}
    \Lcal_G &= \sum_{i=0}^2 \left\{ \Lcal^{cond}_{G_i} + \lambda_{uncond} \Lcal^{uncond}_{G_i} - \lambda_{cycle} \Lcal_{C_i} \right\} + \lambda_{ca}\Lcal_{ca},
\end{align}
where $\lambda_{cycle}$ is the weight of the cycle-consistency term, cross-validated to $\lambda_{cycle}=1.0$. 

\section{Experiments}
\subsection{Dataset and network implementation details}
\label{sec:dataset_details}

Data used in this work was taken from Recipe1M \cite{salvador2017learning}. This dataset contains $\sim$1M recipes with titles, instructions, ingredients and images. We focus on a subset of \num{402760} recipes with at least one image, containing no more than 20 ingredients or instructions, and no less than one ingredient and instruction.
Data is split into 70\% train, 15\% validation and 15\% test sets, using at most \num{5} images from each recipe.  
%
Recipe1M contains $\sim$\num{16}k unique ingredients; we reduce this number by focusing on the 4k most frequent ones. This list is further reduced by first merging the ingredients with the same name after a stemming operation and semi-automatically fusing other ingredients. The later is achieved using a word2vec model trained on Recipe1M, where the ingredients are fused if they are close together in their embedding space and a human annotator accepts the proposed merger.
Finally, we obtain a list of \num{1989} canonical ingredients, covering more than \SI{95}{\%} of all recipes in the dataset.
%
Specific network structures follow those in \cite{chen2018deep} for the association model and \cite{zhang2017stackgan++} for the generator.
The association model is trained on four Tesla K80 for 16 hours (25 epochs) until convergence. 

\subsection{Evaluation of attention-based association model}
\label{sec:comparison_retrieval_models}

To compare with the attention-based model in~\cite{chen2018deep}, we used the setting whose goal is to investigate the ability of the model to, given a query in one modality, retrieve the paired point in the other modality by comparing their similarities in \FoodSpace.  We applied the same metrics as~\cite{chen2018deep}, including the median retrieval rank (MedR) and the recall at top K (R@K). 
MedR is computed as the median rank of the true positive over all queries; the lower MedR 
$\ge$\num{1.0} suggests better performance. 
R@K computes the fraction of true positives recalled among the top-K retrieved candidates; a value between \num{0.0} to \num{100.0} with the higher score indicating better performance.

In \cref{tab:comparison_retrieval}, we report the scores of the attention-based association model \cite{chen2018deep} and our scores with the refined ingredients list. As can be seen, our method decreases MedR from \num{71.0} to \num{24.0} (im2recipe on 5K), clearly showing the advantage of using canonical ingredients instead of the raw ingredients data.
\begin{table}[ht]
\sisetup{round-mode=places,round-precision=3}
    \centering\scriptsize
    \begin{tabular}{llrrrrrrrr}
        \toprule
        && \multicolumn{4}{c}{im2recipe} & \multicolumn{4}{c}{recipe2im} \\ \cmidrule{3-10}
        && MedR$\downarrow$ & R@1$\uparrow$ & R@5$\uparrow$ & R@10$\uparrow$ & MedR$\downarrow$ & R@1$\uparrow$ & R@5$\uparrow$ & R@10$\uparrow$\\
        \midrule
        \multirow{2}{*}{1K} 
        & attention \cite{chen2018deep} &-&-&-&-&-&-&-&- \\ \cmidrule{2-10}
        & ours & \num{5.5} & \num{0.2342} & \num{0.5025} & \num{0.6182} & \num{5.7500} & \num{0.2303} & \num{0.4910} & \num{0.6152} \\
        \midrule
        \multirow{2}{*}{5K} 
        & attention \cite{chen2018deep} & \num{71.0} & \num{0.045} & \num{0.135} & \num{0.202} & \num{70.1} & \num{0.042} & \num{0.133} & \num{0.202} \\ \cmidrule{2-10}
        & ours & \num[math-rm=\mathbf]{24.0} & \num[math-rm=\mathbf]{0.0986} & \num[math-rm=\mathbf]{0.2653} & \num[math-rm=\mathbf]{0.3642} & \num[math-rm=\mathbf]{25.1000} & \num[math-rm=\mathbf]{0.0965} & \num[math-rm=\mathbf]{0.2585} & \num[math-rm=\mathbf]{0.3574} \\
        \midrule
        \multirow{2}{*}{10K} 
        & attention \cite{chen2018deep} & - & - & - & - & - & - & - & - \\ \cmidrule{2-10}
        & ours & \num{47.7} & \num{0.0653} & \num{0.1846} & \num{0.2670} & \num{48.3000} & \num{0.0614} & \num{0.1777} & \num{0.2613} \\
        \bottomrule
    \end{tabular}
    \caption{Comparison with attention-based association model for using image as query to retrieve recipe. `$\downarrow$' means the lower the better, `$\uparrow$' means the higher the better, `-' stands for score not reported in~\cite{chen2018deep}.}
    \label{tab:comparison_retrieval}
\end{table}

\begin{figure*}[ht]
\begin{center}
\includegraphics[width=7cm]{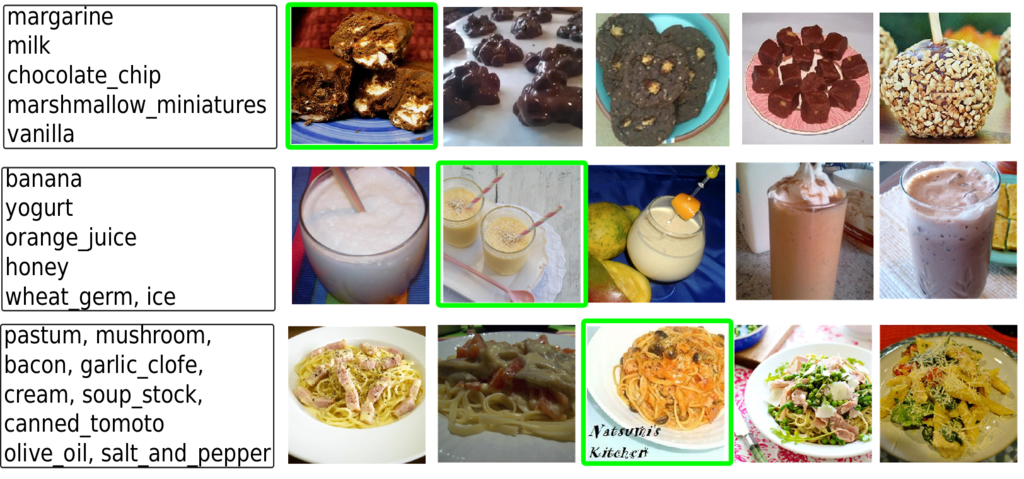}
\end{center}
\caption{Sample results of using ingredients as query to retrieve images on a 1K dataset. Left: query ingredients. Right: top 5 retrieved images (sorted). Corresponding image is indicated by the green box.}
\label{fig:retrieval_examples}
\end{figure*}

\cref{fig:retrieval_examples} illustrates the top 5 retrieved images using the ingredients as the query.
Although the correct images do not always appear in the first position, the retrieved images largely belong to the same food type, suggesting commonality in ingredients.

\vspace{-0.5cm}
\subsection{Meal Image Synthesis}
In this section we present the results of synthesizing meal image given an ingredient set. 
To mitigate the diversity caused by different preparation methods, we focus on narrow meal categories where the cutting and cooking methods are largely consistent within each category. 
In the following experiments, we only train on specific types of food within three commonly-seen categories: salad, cookie, and muffin. 
Images from these categories usually contain key ingredients that are easily recognized, which can be used to verify the model's ability to manipulate meal image by changing those ingredients. 
The number of samples in train/test dataset are $17209/3784$ (salad), $9546/2063$ (cookie) and $4312/900$ (muffin).

Evaluating the performance of generative models is generally a challenging task.
In this paper, we choose inception score (IS) \cite{salimans2016improved} and Frechet inception distance (FID) \cite{heusel2017gans} as our quantitative evaluation metrics.
%
Intuitively, a good model should generate images that are both meaningful
and diverse,
which would correspond to a high IS.
The drawback of IS is that the inception model is trained on ImageNet; those image features may not be very reflective of meal images.
FID assumes the image features follow a Gaussian distribution and measures the Frechet distance between the real and the synthesized data distributions. 
Intuitively, low FID means the two distributions are close. FID shares some of the same drawbacks as IS, although to somewhat lesser extent.

\begin{figure*}[t]
\begin{center}
\includegraphics[width=\textwidth]{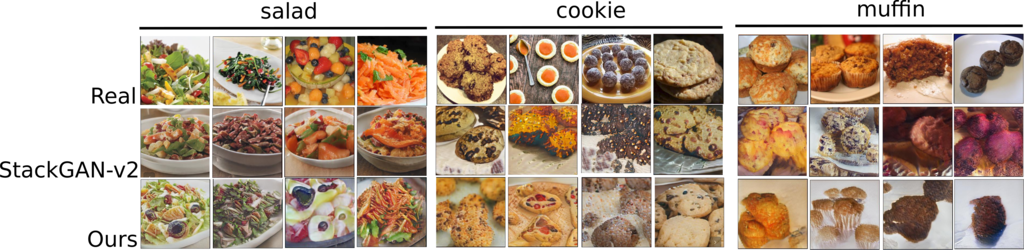}
\end{center}
\caption{Example results from different ingredients with same random vector.}
\label{fig:different_ingredients_same_noises}
\end{figure*}

We compute IS and FID on \num{900} samples randomly generated on the test-set for each category, which is the maximum number of recipes among test-sets for salad, cookie and muffin. 
The IS of real images are also computed as a baseline.
\cref{tab:is_and_fid} shows the results obtained on different categories.
Our model achieves better IS and FID on most subsets.
\begin{table}[ht]
\sisetup{round-mode=places,round-precision=2, separate-uncertainty}

\begin{tabular}[t]{cc}

\begin{subtable}[t]{0.48\textwidth}

    \centering\scriptsize
    \begin{tabular}[t]{clrrr}
        \toprule
         & & salad & cookie & muffin \\
        \midrule
        \multirow{3}{*}{IS $\uparrow$}  & StackGAN-v2 & \num{3.0652} & \num[math-rm=\mathbf]{4.6992}  & \num{2.6020} \\ \cmidrule{2-5}
                                        & ours & \num[math-rm=\mathbf]{3.4573} & \num{2.8172} & \num[math-rm=\mathbf]{2.9368} \\ \cmidrule{2-5}
                                        & real & \num{5.1205} & \num{5.6989} & \num{4.1968} \\
        \midrule
        \multirow{2}{*}{FID $\downarrow$}   & StackGAN-v2 & \num[math-rm=\mathbf]{55.4263} & \num{106.1381} & \num{104.7341} \\ \cmidrule{2-5}
                                            & ours & \num{78.7922} & \num[math-rm=\mathbf]{87.1371} & \num[math-rm=\mathbf]{81.1320} \\
        \bottomrule
    \end{tabular}
    \caption{Inception score (IS) and Frechet inception distance (FID).}
    \label{tab:is_and_fid}
    
\end{subtable}

&
     
\begin{subtable}[t]{0.48\textwidth}

    \centering\scriptsize
    \begin{tabular}[t]{lrrr}
        \toprule
         & salad & cookie & muffin \\
         \midrule
        random      & \num{450.0}   & \num{450.0}   & \num{450.0}   \\
        StackGAN-v2  & \num[math-rm=\mathbf]{58.4000}  & \num{194.45}  & \num{217.5000}  \\
        ours        & \num{66.1500}   & \num[math-rm=\mathbf]{103.300}   & \num[math-rm=\mathbf]{211.0000}   \\
        real        & \num{12.15}   & \num{47.35}   & \num{65.0}    \\
        \bottomrule
    \end{tabular}
    \caption{Median rank comparison.}
    \label{tab:comparison_retrieval_synthesis}
    
\end{subtable}

\end{tabular}

    \caption{Performance analysis: (a) Comparison of StackGAN-v2 and our model on different subsets by inception scores (IS) and Frechet inception distance (FID). (b) Comparison of median rank (MedR) by using synthesized images to retrieve recipes in subsets. We choose $900$ as the retrieval range
    to adhere to the maximum number of recipes among test-sets for salad, cookie and muffin. 
    }

\end{table}
\cref{fig:different_ingredients_same_noises} shows examples generated from different ingredients with the same random vector $\vec{z}$. 
Within each category, the generated images capture the main ingredients for different recipes while sharing a similar view point. This demonstrates the model's ability to synthesize meal images conditioned on ingredient features $\vec{c}$ while keeping nuisance factors fixed through vector $\vec{z}$.
Compared with StackGAN-v2, the images generated by our model appear to contain ingredients that are more like those in the real image.
In the third column of salad, \eg, the ingredients in our image are more like fruit.

\begin{figure*}[t]
\begin{center}
\includegraphics[width=\textwidth]{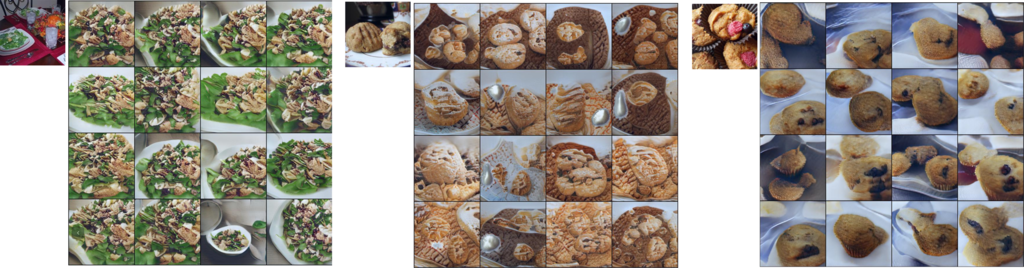}
\end{center}
\caption{Example results from same ingredients with different random vectors. \num{16} synthesized images are shown for each real image (top-left).}
\label{fig:same_ingredients_different_noises}
\end{figure*}

\cref{fig:same_ingredients_different_noises} further demonstrates the different roles of ingredients appearance $\vec{c}$ and random vector $\vec{z}$ by showing examples generated from same ingredients with different random vectors. 
The synthesized images have different view points, but still all appear to share the same ingredients.

\begin{figure*}[t]
\begin{center}
\includegraphics[width=12cm]{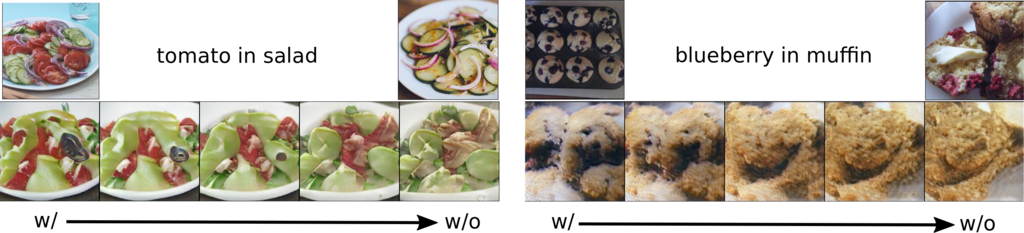}
\end{center}
\caption{Example results of synthesized images from the linear interpolations in \FoodSpace between two recipes (with and without target ingredient). Target ingredient on the left is tomato and the model is trained with salad subset; target ingredient on the right is blueberry and the model is trained with muffin subset. The interpolation points from left to right are $\frac{with}{without} = \left\{ \frac{4}{0}, \frac{3}{1}. \frac{2}{2}, \frac{1}{3}, \frac{0}{4} \right\}$}
\label{fig:linear_interpolations}
\end{figure*}

To demonstrate the ability to synthesize meal images corresponding to specific key ingredient, we choose a target ingredient and show the synthesized images of linear interpolations between a pair of ingredient lists $r_i$ and $r_j$ (in the feature space), in which $r_i$ contains the target ingredient and $r_j$ is without it, but shares at least \SI{70}{\%} of remaining ingredients in common with $r_i$\footnote{The reason for choosing the partial overlap is because very few recipes differ in exactly one key ingredient.}. One can observe that the model gradually removes the target ingredient during the interpolation-based removal process, as seen in \cref{fig:linear_interpolations}.

We also investigate the median rank (MedR) by using synthesized images as the query to retrieve recipes with the association model in \cref{sec:comparison_retrieval_models}. \cref{tab:comparison_retrieval_synthesis} suggests our method outperforms StackGAN-v2 on most subsets, indicating both the utility of the ingredient cycle-consistency and the embedding model. Still, the generated images remain apart from the real images in their retrieval ability, affirming the extreme difficulty of the photo-realistic meal image synthesis task.

\section{Conclusion}
In this paper, we develop a model for generating photo-realistic meal images based on sets of ingredients. 
We integrate the attention-based recipe association model with StackGAN-v2, aiming for the association model to yield the ingredients feature close to the real meal image in \FoodSpace, with StackGAN-v2 attempting to reproduce this image class from the \FoodSpace encoding. To improve the quality of generated images, we reuse the image encoder in the association model and design an ingredient cycle-consistency regularization term in the shared space. Finally, we demonstrate that processing the ingredients into a canonical vocabulary is a critical key step in the synthesis process.
Experimental results demonstrate that our model is able to synthesize natural-looking meal images corresponding to desired ingredients, both visually and quantitatively, through retrieval metrics.
In the future, we aim at adding additional information including recipe instructions and titles to further contextualize the factors such as the meal preparation, as well as combining the amount of each ingredient to synthesize images with arbitrary ingredients quantities.

\bibliography{egbib}
\end{document}